%% file: main.tex
\documentclass[10pt,twocolumn,letterpaper]{article}

\usepackage{xcolor}
\usepackage[pagenumbers]{arxiv} 


\definecolor{ourblue}{rgb}{0.21,0.49,0.74}
\usepackage[pagebackref,breaklinks,colorlinks,allcolors=ourblue]{hyperref}

\usepackage{multirow}

\title{CRoF: CLIP-based Robust Few-shot Learning on Noisy Labels}
\author{Shizhuo Deng$^{1}$\thanks{Equal contribution.}, Bowen Han$^{1}$\footnotemark[1],  Jiaqi Chen$^{1}$, Hao Wang$^{1}$,  Dongyue Chen$^{1}$\thanks{Corresponding author.}, Tong Jia$^{1}$\\
$^{1}$ College of Information Science and Engineering,\\
Northeastern University Shenyang 110819, Liaoning, China\\
{\tt\small \{dengshizhuo, chendongyue\}@ise.neu.edu.cn}\\
}

\begin{document}
\maketitle
\input{sec/0_abstract}    
\input{sec/1_intro}
\input{sec/2_related}
\input{sec/3_method}

\input{sec/4_ex}

\input{sec/5_conclusion}

{
    \small
    \bibliographystyle{ieeenat_fullname}
    \bibliography{main}
}


\end{document}

%% file: sec/0_abstract.tex
\begin{abstract}

Noisy labels threaten the robustness of few-shot learning (FSL) due to the inexact features in a new domain. CLIP \cite{radford2021learning}, a large-scale vision-language model, performs well in FSL on image-text embedding similarities, but it is susceptible to misclassification caused by noisy labels. How to enhance domain generalization of CLIP on noisy data within FSL tasks is a critical challenge. In this paper, we provide a novel view to mitigate the influence of noisy labels, \textbf{C}LIP-based  \textbf{Ro}bust \textbf{F}ew-shot learning (CRoF). CRoF is a general plug-in module for  CLIP-based models. To avoid misclassification and confused label embedding, we design the few-shot task-oriented prompt generator to give more discriminative descriptions of each category. The proposed prompt achieves larger distances of inter-class textual embedding. Furthermore, rather than fully trusting zero-shot classification by CLIP, we fine-tune CLIP  on noisy few-shot data in a new domain with a weighting strategy like label-smooth. The weights for multiple potentially correct labels consider the relationship between CLIP’s prior knowledge and original label information to ensure reliability. Our multiple label loss function further supports robust training under this paradigm. Comprehensive experiments show that CRoF,  as a plug-in, outperforms fine-tuned and vanilla CLIP models on different noise types and noise ratios.



\end{abstract}





%% file: sec/1_intro.tex
\section{Introduction}

Low-quantity data, such as noisy labels, has a great negative impact on feature space and decision boundaries in deep learning models, especially in the few-shot learning (FSL) tasks.
Current FSL models typically assume clean data, which is unrealistic in real-world applications\cite{Liang_2022_CVPR}.
Incorrect labels are inevitable due to manual labeling, semi-automated labeling, and so on.
The limited number of labeled samples makes FSL highly sensitive to noisy labels.
Therefore, learning from noisy labels is becoming a practical and urgent problem for robust few-shot learning.

\begin{figure}[t]
  \centering
  \includegraphics[width=1\linewidth]{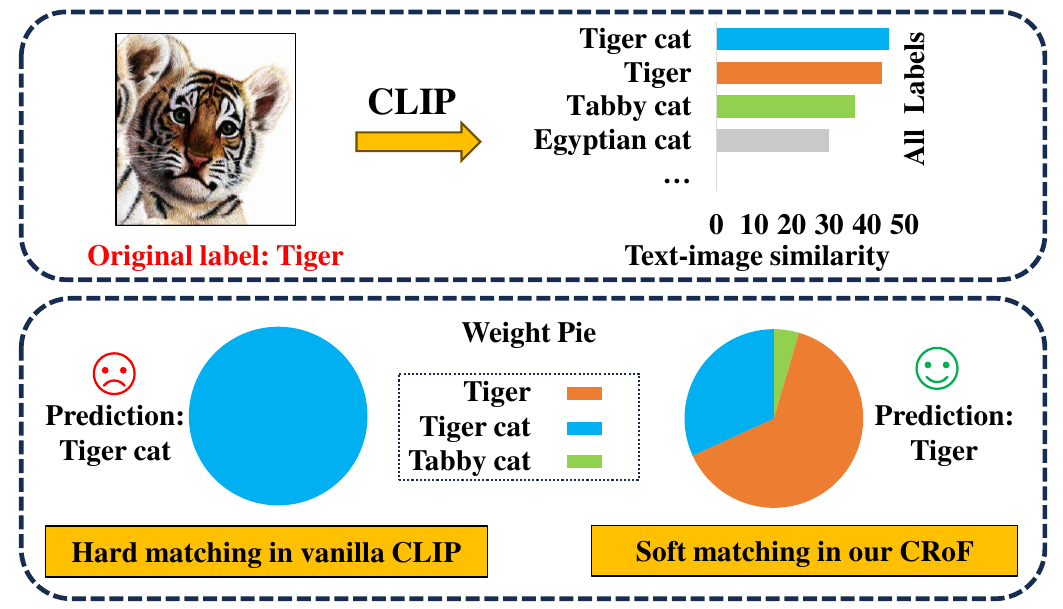}
  \caption{Example of misclassification by CLIP and comparison on different matching methods.}
  \label{fig:intro}
\end{figure}

Typical few-shot learning on noisy labels focuses on refined category prototypes  \cite{Liang_2022_CVPR, Mazumder_2021_WACV} and reducing the influence of incorrect labels \cite{9072304} on only image modality.
They learn the image embedding distribution in feature space to avoid the noisy sample's negative effect.
However, single-modal limits the upper bound of generalization, and the traditional FSL is usually constrained to fixed and small class sets (e.g., 5-way).
It is unsuitable for scenarios with many classes but few samples per class.
Meanwhile,  multi-modal information, such as image-text pairs,  has been investigated for robust learning under noisy labels  \cite{NEURIPS2021_f5e62af8, Han_2023_CVPR, Yang_2024_CVPR}.
They utilize metric learning to correct the noisy correspondence on similarities between image-text pairs, such as triplet loss.
However, aligning image-text features requires extensive training data, which makes these approaches impractical for FSL where labeled samples are scarce.
Thus, there remains a gap in robust few-shot tasks with multi-modal data. 

To bridge the above gap,  CLIP \cite{radford2021learning}, a pre-trained large-scale vision-language model on 400 million image-text pairs, is a good choice to obtain the similarity.
CLIP enhances the performance in few-shot learning.
However, CLIP-based methods also encounter the problem of mismatching when handling noisy labels in few-shot learning, similar to the previous studies \cite{NEURIPS2021_f5e62af8, Han_2023_CVPR, Yang_2024_CVPR}.
For instance, in \cref{fig:intro}, vanilla CLIP may provide the incorrect prediction of "Tiger cat" with the highest similarity.
The original label "Tiger"  ranks second in the top three highest similarity label set, which is a confusing result.
Based on whether the original labels are ground truth or not, we discuss two scenarios. 
\textbf{(1)} The original label is the ground truth. The incorrect matching is conducted by CLIP with noisy correspondence.
\textbf{(2)} The most similar label calculated by CLIP is the ground truth. It means the original label is wrong. 
Considering the noisy label and noisy correspondence, it is difficult to judge the exact labels.

Therefore, achieving a trade-off between the original label and the most similar label produced by CLIP is a great challenge, which is crucial for enhancing the accuracy of CLIP-based models under noisy few-shot conditions. 
There are two questions to be solved.

\textbf{Q1: How to increase the distance of the embedding of the textual label description?} 
Although the category label embedding from CLIP's text encoder can achieve good classification, the embedding similarity between similar categories remains high. 
Noisy label exacerbates this situation, negatively impacting text-image similarity and increasing the likelihood of misclassification.

\textbf{Q2: How to balance the contributions of the original labels and the produced labels by CLIP for robust FSL?}
Hard matching with most similar label has a high risk in noisy settings, as shown in  \cref{fig:intro}.
In contrast, soft matching helps retain potential ground truth labels.
So, the reasonable weights of the candidate labels enhance the robustness, which is similar to the label smoothing mechanism \cite{Szegedy_2016_CVPR,muller2019does}.

Therefore, we propose \textbf{C}LIP-based  \textbf{Ro}bust \textbf{F}ew-shot learning, abbreviated as CRoF, a plug-in solution aimed at enhancing data quality for robust FSL. 
For one thing, appropriate prompts are designed to distinguish the embedding of the category description, which is crucial in sequential text-image matching.
For another, we fine-tune CLIP on a noisy few-shot dataset in a new domain and assign weights to the candidate labels on the similarity ranking list, rather than giving complete confidence to CLIP matching results and the original label.
Our main contributions are as follows:

\begin{itemize}

   \item We present a novel perspective to weaken the impact of the noisy label in few-shot learning tasks by using a large-scale vision-language model. The proposed CRoF serves as a plug-in solution for all CLIP-based few-shot models.
   
  \item The task-oriented prompt generator is designed to align the label descriptions with the target. The supplement information is an augmented prompt to basic prompts, which helps decrease the similarity among label embeddings.

    \item We propose a multiple label weighting strategy that combines the power of CLIP's prior knowledge and original label information, reducing the risk of missing the ground-truth label.

   \item  We conduct comprehensive experiments on various public datasets with different noise settings to validate the effectiveness of CRoF. CRoF demonstrates strong robustness and is less sensitive to noisy labels.
\end{itemize}

%% file: sec/2_related.tex
\section{Related Works}
\subsection{Few-shot Learning on Noisy Labels} \label{sec:re1}
Traditional few-shot learning \cite{snell2017prototypical, vinyals2016matching, sung2018learning, finn2017model} encounters the problem of noisy labels. 
The existing methods address the issue by constructing more robust prototypes and weighting labels based on reliability.
RNNP \cite{Mazumder_2021_WACV} generates mixed prototypes through linear interpolation of samples in the support set, to enhance the robustness of prototypes.
TraNSF \cite{Liang_2022_CVPR} addresses the limitations of few-shot models in handling noisy labels by replacing the mean operator used in ProtoNet \cite{snell2017prototypical} with a more robust aggregation method. 
To calculate label weights based on reliability, RapNet \cite{9072304} designs a parametric attention module by a learnable bidirectional long short-term memory network \cite{6795963} and a fully connected network \cite{Long_2015_CVPR}, which generates attention weights to effectively weaken noisy labels in few-shot learning scenarios.  
DCML \cite{que2024dual} utilizes meta-learning with class-level sampling and example-level sampling to dynamically adjust during training, minimizing the impact of noisy labels.
Besides, other data structure, such as graph, is utilized for robust learning on few-shot settings \cite{chen2024appn}.

\begin{figure*}[ht]
  \centering
  \includegraphics[width=0.95\linewidth]{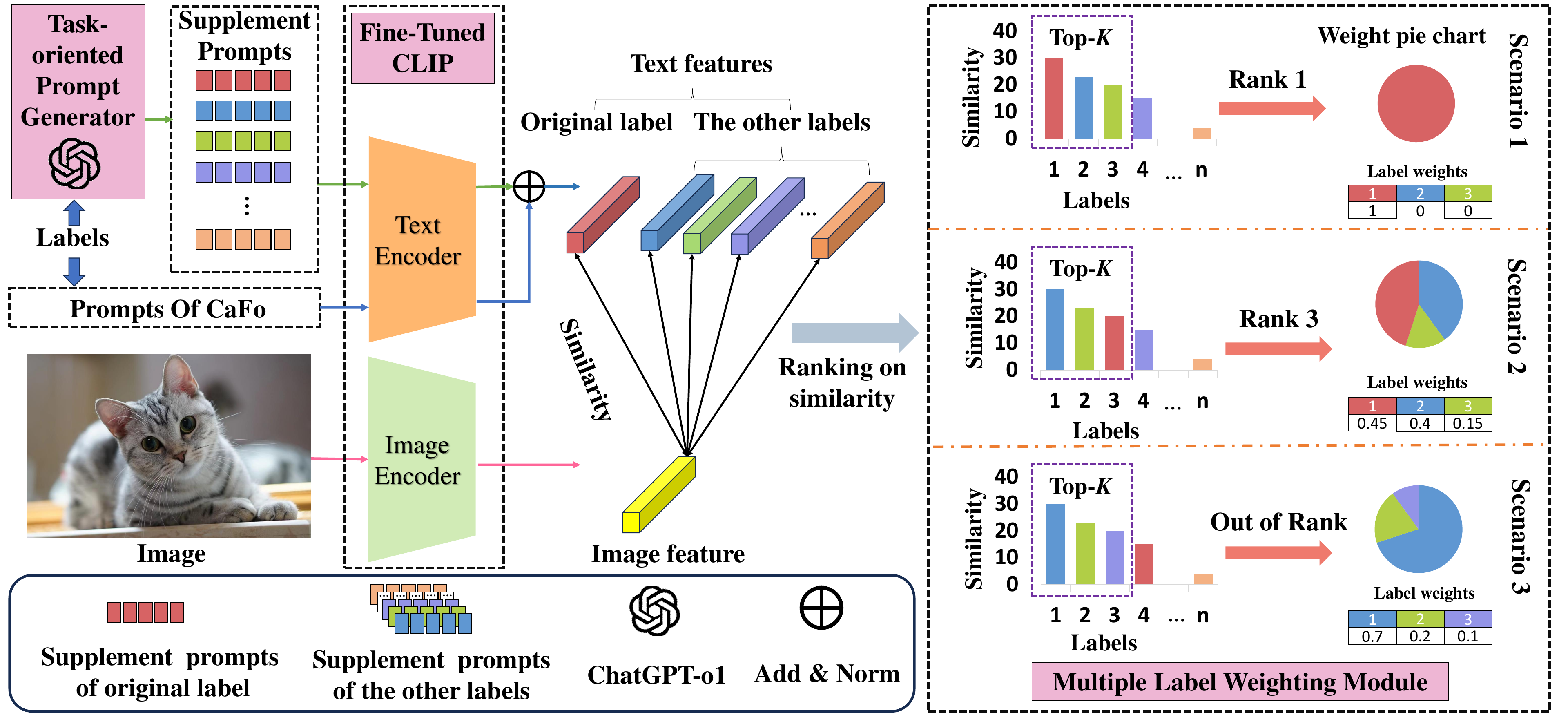}
  \caption{Overview of CRoF. There are three components in CRoF, task-oriented prompt generator, fine-tuned CLIP, and multiple label weighting module.
  The red bar represents the similarity between the original label text feature and its image feature. According to the position of the red bar in the \textit{top-K} similarities, there are three scenarios.
  }
    \label{fig:overall}
\end{figure*}

\subsection{Few-shot Learning with CLIP} Fine-tuned CLIP \cite{radford2021learning}, which captures more accurate image-text relationships through cosine similarity on few-shot datasets, can be divided into three main approaches.
\textbf{(1) Prompt tuning.} CoOp \cite{zhou2022coop} and CoCoOp \cite{zhou2022cocoop} enhance few-shot learning by transforming CLIP's text descriptions ("a photo of [class]") into learnable vectors. While these methods achieve excellent results, they require more computational resources and training time.
\textbf{(2) Adapter tuning.} CLIP-Adapter \cite{gao2024clip} offers a lightweight fine-tuning approach by adding two linear layers directly after the image encoder, preserving CLIP's capabilities with residual connections to prevent forgetting CLIP's zero-shot knowledge. 
\textbf{(3) Cache tuning.} Tip-Adapter \cite{zhang2022tip} constructs a cache model using image features and labels from the training set, presenting a training-free CLIP fine-tuning method to better fit on few-shot datasets. Tip-Adapter-F further enhances adaptation by unfreezing the cache model with the adapter method.
CaFo \cite{Zhang_2023_CVPR} combines these methods by using GPT-3 \cite{brown2020language} for text prompts and DALL-E \cite{pmlr-v139-ramesh21a} for image generation.  Cache-based models combine the textual and visual prior knowledge of CLIP and DINO \cite{Caron_2021_ICCV} to achieve better few-shot learning results. 

Recently, CLIP-based few-shot learning methods \cite{Tang_2024_CVPR, Silva-Rodriguez_2024_CVPR, Huang_2024_CVPR} have become more popular.
However, they have not considered the potential negative impact of fine-tuning on noisy datasets. Robust few-shot learning with CLIP is promising and valuable.

%% file: sec/3_method.tex

\section{Method}
\subsection{Overview}
To address noisy labels and the noisy correspondence in few-shot learning tasks, we propose \textbf{C}LIP-based  \textbf{Ro}bust \textbf{F}ew-shot learning, (CRoF), comprising three components, as shown in  \cref{fig:overall}.
The task-oriented prompt generator provides target prompts with smaller inter-class similarity, facilitating the distinction of the similar classes.
Fine-tuned CLIP  improves the matching accuracy of CLIP in few-shot learning tasks in a new domain. 
In the multiple label weighting module, the hard matching mechanism in the vanilla CLIP is expanded to  \textit{top-K} most similar labels with appropriate weights, which aims to cover the right labels and improve the robustness in label correction.

\subsection{Task-oriented Prompt Generator}
To reduce the similarity among categories and give supplement prompts, we design the following ChatGPT requests.

\textit{I have \textbf{n} categories of [task targets (such as car, action, flower, etc.)], and each category is described as "a photo of {category name}." I want to introduce \textbf{detailed differentiation descriptions, comparative information, scene backgrounds, emotions, or domain-specific terms} in the descriptions to guide  CLIP text encoder to generate more distinguishable category embedding for similar categories. Please generate five descriptions for each category according to above principles. My category list is: [category-1, …, category-n]. Output the descriptions in JSON format.}

\begin{figure}[ht]
  \centering
  \includegraphics[width=0.9\columnwidth]{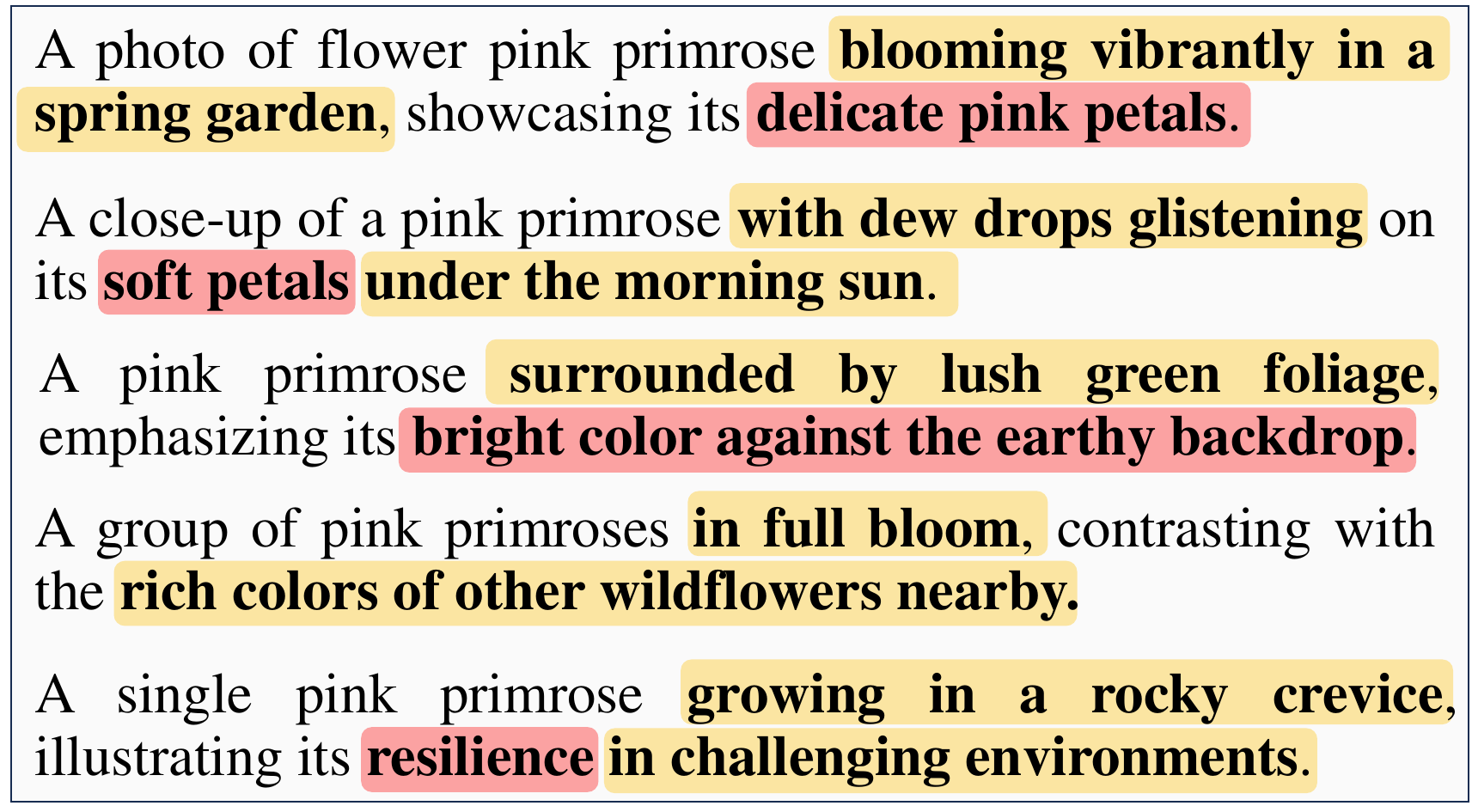}
  \caption{An example of  generated description of pink primrose.}
    \label{fig:random}
\end{figure}

The generated descriptions for  \textit{pink primrose} are shown in \cref{fig:random}, where the yellow part indicates supplement details and the red part is the original corresponding information of the target category.
The yellow part provides more randomness under category semantic constraints. 
The red part shows descriptions that are highly related to category information.
Our descriptions generate the supplement prompts that retain the original information while also increasing inter-class differentiation.

To maintain the image-text feature consistency and avoid conflicts from the CLIP's prior knowledge, we treat the randomness of supplement details as perturbed information to CaFo \cite{Zhang_2023_CVPR}.
CaFo generates multiple descriptions for each category in a question-and-answer format.
We input both prompts into the CLIP text encoder, allowing the augmented prompts to increase the inter-class distances in label description embedding.
Let  $E^T$ = $\{ e^T_1, e^T_2, \ldots, e^T_n \}$ = $\{ f_e(l^T_1), f_e(l^T_2), \ldots, f_e(l^T_n) \}$  be the supplement description embedding and  $E^C$ = $\{ e^C_1, e^C_2, \ldots, e^C_n \} $=$ \{ f_e(l^C_1), f_e(l^C_2), \ldots, f_e(l^C_n) \}$ is  CaFo description embedding, where $f_e()$ is the CLIP text encoder, $l_i$ is the description.
The embedding of the task-oriented prompt generator (TPG) is $E$=$\{ e_i=Norm (e^T_i+e^C_i)|i=1,2,\dots, n\}$.

\subsection{Fine-Tuned CLIP}
\begin{figure}[ht]
  \centering
  \includegraphics[width=0.9\columnwidth]{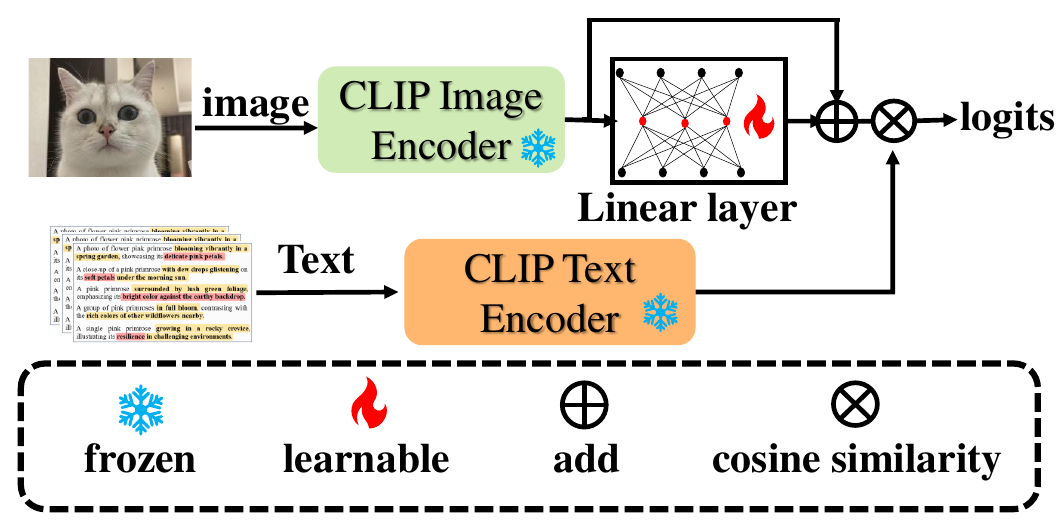}
  \caption{Structure of CLIP-Adapter.}
    \label{fig:clip}
\end{figure}

To improve CLIP's accuracy in few-shot tasks and detect noisy correspondence, we use the CLIP-Adapter \cite{gao2024clip} to fine-tune CLIP in \cref{fig:clip}. 
The CLIP text encoder \begin{math}f_e()\end{math} and CLIP image encoder \begin{math}f_m()\end{math}  map labels and images into an aligned feature space.
The image feature  \begin{math}X\end{math} is fine-tuned through an adapter network consisting of two linear layers. In order not to forget the powerful ability of the original CLIP, the image feature is adjusted using residual structure. The fine-tuned image feature can be calculated by \cref{eq:image}, where \begin{math}W_1\end{math} and \begin{math}W_2\end{math} represent the learnable parameters of the two linear layers, the residual parameter is $\lambda$.
\begin{equation}\label{eq:image}
  \begin{split}
    X^* &= (1-\lambda)X + \lambda \textit{adapter}(X) \\
        &= (1-\lambda)X + \lambda(\textit{ReLU}(XW_1)W_2),
  \end{split}
\end{equation}

The similarity score 
\begin{math}
S = \{ s_1, s_2, \ldots, s_n \} 
\end{math} 
and the probability of  each category 
\begin{math}
P = \{ p_1, p_2, \ldots, p_n \} 
\end{math} 
are calculated through the fine-tuned image feature in \cref{eq:si} and \cref{eq:pi}, where $\cos()$ is cosine similarity and $\tau$ is a temperature parameter for softmax.
\begin{equation}\label{eq:si}
  s_i = \exp\left({\text{cos}(X^*, e_i)}/{\tau}\right),
\end{equation}
\begin{equation}\label{eq:pi}
  p_i = \frac{s_i} {\sum_{j=1}^n {s_j}} = \frac{{\exp\left({\text{cos}(X^*, e_i)}/{\tau}\right)}} {\sum_{j=1}^n {\exp\left({\text{cos}(X^*, e_j)}/{\tau}\right)}},
\end{equation}

During fine-tuning, the two linear layers can be optimized using the cross-entropy loss in \cref{eq:l1}.
 $M$ is the total number of training samples. 
$\theta = \{ W_1, W_2 \}$ represents all learnable parameters.
\begin{equation}\label{eq:l1}
  L(\theta) = -\frac{1}{M} \sum_{m=1}^{M} \sum_{i=1}^{n} y^{(m)}_i \log p^{(m)}_i,
\end{equation}\textbf{}

\subsection{Multiple Label Weighting Module}

We replace the hard matching in CLIP with soft matching mechanism, employing multiple label smoothing to reduce the risk of missing the ground-truth label.
We design a weighting strategy to assign weights to the possible correct labels with  \textit{top-K}  highest similarity while training.

Our goal is to achieve a balance between the most similar label and the original label, applying label smoothing across all candidate labels.
As discussed in \cref{fig:intro}, we aim for the original label to have a relatively high weight, \cite{Szegedy_2016_CVPR} but still be constrained by its ranking. 
The weight of \textit{top}-1 label generated by CLIP is smaller than that of the original label before a certain ranking.
The weights of the remaining labels are determined on the distance from the \textit{top}-1 label.

Similarities calculated by \cref{eq:si} are sorted in descending order in $S^*$=$\{ s_1^*,  \ldots ,s_n^*\}$.
Note that, the similarity between the original label text-image pair is  $s_r^*$, $s_r^*\in S^*$, where  $r$ represents the rank of  \( s_r^* \) in \( S^* \), 
\( 1 \leq r \leq n \).
\textit{Top-K} most similar labels are denoted as 
$S_K^*$=$\{s_1^*,\ldots,s_K^*\}$.
We calculate the weights $w$=$\{w_1, \ldots, w_K\}$ for $K$ candidate labels based on their ranking position $r$ in three scenarios.

\textbf{(1) Scenario 1: \( \mathbf{\textit{r} = 1} \).}
When the original label ranks first, the weights are one-hot format.
The original label is supposed to be correct. The weight of the original label is 1, and the weights of the other  \( k-1 \) labels are set to 0. 


\textbf{(2) Scenario 2: \( \mathbf{1 < \textit{r} \leq \textit{K}} \).}
Noisy annotation and noisy correspondence lead to this scenario.
Both the original label and the most similar label are valuable.
Considering the distribution of ground truth, we set  $\alpha$ to balance the original label and the most similar label, where $\alpha$ represents the degree of loyalty to the original label and \( \alpha \in (0,1) \). As the rank \( r \) increases, the weight should decrease, which is \( \alpha \cdot \gamma^{r-2} \), \( \gamma \in (0,1) \) in \cref{eq:2}.

    \begin{equation}
        w_i = 
        \begin{cases}
            (1-\alpha \cdot \gamma^{r-2}) \cdot \beta & \textit{if } i = 1, \\[10pt]
            \alpha \cdot \gamma^{r-2} & \textit{if } i = r, \\[10pt]
            (1-\alpha \cdot \gamma^{r-2}) \cdot (1-\beta) \\ \cdot \dfrac{s_1^* - s_i^*}{\textstyle \sum_{j=2 \wedge j\neq r}^{k}(s_1^* - s_j^*)} & \textit{otherwise}.
        \end{cases}
        \label{eq:2}
    \end{equation}

The remaining weight,  1 - $\alpha \cdot \gamma^{h-2}$, is distributed  into  the   $K$-1 labels. 
Unlike the uniform distribution \cite{Szegedy_2016_CVPR} for the remaining labels, our approach leverages a pre-trained large-scale vision-language model, so the label with the highest similarity should naturally receive a larger weight.
However,  too much weight may ignore the contribution of possible correct information from the remaining $K$-2  labels. Therefore, we introduce another hyperparameter \( \beta \) to achieve a trade-off for the most similar label.
 At this stage, $w_1 $ = \( \beta \cdot (1 - \alpha \cdot \gamma^{h-2}) \), and the weights of the remaining  $K$-2  labels are allocated based on their distances from the label with the highest similarity, denoted as $s_1^*$-$  s_i^*$. The weight $w$ can be calculated by \cref{eq:2}.

\textbf{(3) Scenario 3: \( \mathbf{\textit{K} < \textit{r} \leq \textit{n}} \).}
If the rank $r$ is out of the range of \textit{K}, the original label is not considered to be ground truth and is discarded. 
The weights of \textit{top-K} labels are determined by hyperparameter \( \beta \) to balance the most similar label and others. The weights of the remaining \( K-1 \) labels follow the rules of \cref{eq:2} and  calculated by
    \begin{equation}
        w_i = 
        \begin{cases} 
          \beta & \textit{if } i = 1, \\[10pt]
          (1 - \beta) \cdot \dfrac{s_1^* - s_i^*}{\sum\limits_{j=2}^{k}(s_1^* - s_j^*)} & \textit{otherwise}.
        \end{cases}
        \label{eq:3}
    \end{equation}

To integrate the relationship of text-image pairs with the smoothed weights, we normalize the $w$ as $w^*$= $\{w^*_1, \ldots, w^*_K\}$ as shown in \cref{eq:7}.

\begin{equation}\label{eq:7}
  w_i^* = \frac{s_i^*.w_i}{\sum\limits_{j=1}^{k}s_j^*.w_j}.
\end{equation}

\subsection{Overall Objective}
Instead of the loss function on a one-hot format, 
the overall objective function is defined in \cref{eq:8}, which consists of $K$ original base loss function $L_{mk}^{O}$ on the hard label.
Therefore, the multiple label weighting can be used as a plug-in module in any CLIP-based model.
For example, $L_{mk}^{O}$ could be cross-entropy loss of one sample, $ -y^{(m)}_k \log p^{(m)}_k$, within \textit{top-K} candidate set.

\begin{equation}\label{eq:8}
  L = \frac{1}{M} \sum_{m=1}^{M} \sum_{k=1}^{K}L_{mk}^{O}.w_{mk}^*
\end{equation}

%% file: sec/4_ex.tex
\section{Experiments}
\subsection{Experimental Settings}
\textbf{(1) Datasets and Noisy Label Setting}

The diversity of the evaluation dataset ensures a comprehensive evaluation of our method across various types of image recognition tasks.
Primary experiments are conducted on the public datasets, general object recognition dataset \textit{Caltech101} \cite{1384978} and the action recognition dataset \textit{UCF101} \cite{soomro2012ucf101dataset101human}. 
 Generalization is verified on the other nine datasets in four tasks: generic object recognition (\textit{ImageNet} \cite{5206848}, \textit{miniImageNet}\cite{vinyals2016matching}, \textit{tieredImageNet}\cite{46640}), fine-grained image recognition  (\textit{OxfordPets} \cite{6248092}, \textit{StanfordCars} \cite{Krause_2013_ICCV_Workshops}, \textit{Flowers102} \cite{4756141}, and \textit{Food101} \cite{10.1007/978-3-319-10599-4_29}), texture classification (\textit{DTD} \cite{Cimpoi_2014_CVPR}) and scene recognition (\textit{SUN397} \cite{5539970}). 
We simulate the few-shot learning task by limiting the number of samples per class, 5-shot and 10-shot.
However, increasing the number of categories is encouraged to enhance classification difficulty, different from the limited categories in traditional few-shot.

\begin{table*}[ht]
    \centering
    \setlength{\tabcolsep}{1mm}
    \begin{tabular}{cccccccccccc}
        \toprule
        \multirow{2}{*}{datasets} & \multirow{2}{*}{methods} & \multicolumn{5}{c}{5-shot} & \multicolumn{5}{c}{10-shot} \\ \cmidrule(lr){3-7} \cmidrule(lr){8-12}
         &  & 0 & 0.2 & 0.4 & 0.6 & 0.8 & 0 & 0.2 & 0.4 & 0.6 & 0.8 \\ \midrule
        \multirow{4}{*}{\textit{UCF101}} & CLIP-Adapter & 72.09 & 66.35 & 63.2 & 60.22 & 50.81 & 73.43 & 66.88 & 63.65 & 60.96 & 52.42 \\ 
         & (CLIP-Adapter)+CRoF & 72.4$^{\uparrow}$ & 69.44$^{\uparrow}$ & 67.59$^{\uparrow}$ & 62.52$^{\uparrow}$ & 57.39$^{\uparrow}$ & 74.89$^{\uparrow}$ & 72.64$^{\uparrow}$ & 69.87$^{\uparrow}$ & 65.79$^{\uparrow}$ & 57.41$^{\uparrow}$ \\ 
         & Tip-Adapter-F & 74.2 & 70.34 & 64.26 & 56.89 & 50.83 & 75.5 & 67.86 & 61.62 & 57.1 & 45.47 \\ 
         & (Tip-Adapter-F)+CRoF & 74.31$^{\uparrow}$ & 70.24$^{\downarrow}$ & 64.68$^{\uparrow}$ & 60.32$^{\uparrow}$ & 52.92$^{\uparrow}$ & 75.31$^{\downarrow}$ & 71.85$^{\uparrow}$ & 68.38$^{\uparrow}$ & 62.07$^{\uparrow}$ & 53.74$^{\uparrow}$ \\ 
        \midrule
        \multirow{4}{*}{\textit{Caltech101}} & CLIP-Adapter & 90.83 & 89.05 & 87.99 & 84.34 & 79.96 & 91.81 & 90.43 & 90.18 & 88.32 & 82.39 \\ 
         & (CLIP-Adapter)+CRoF & 91.48$^{\uparrow}$ & 90.63$^{\uparrow}$ & 90.47$^{\uparrow}$ & 89.13$^{\uparrow}$ & 85.48$^{\uparrow}$ & 91.36$^{\downarrow}$ & 90.99$^{\uparrow}$ & 90.47$^{\uparrow}$ & 89.82$^{\uparrow}$ & 85.6$^{\uparrow}$ \\ 
         & Tip-Adapter-F & 90.87 & 89.13 & 87.1 & 73.71 & 62.39 & 91.89 & 88.19 & 84.62 & 73.59 & 53.43 \\ 
         & (Tip-Adapter-F)+CRoF & 90.91$^{\uparrow}$ & 89.94$^{\uparrow}$ & 89.01$^{\uparrow}$ & 84.18$^{\uparrow}$ & 80.28$^{\uparrow}$ & 91.97$^{\uparrow}$ & 89.41$^{\uparrow}$ & 87.1$^{\uparrow}$ & 85.56$^{\uparrow}$ & 77.77$^{\uparrow}$ \\ 
        \bottomrule
    \end{tabular}
    \caption{Accuracy (\%) on symmetric noise.}
    \label{tab:symmetric}
\end{table*}

\begin{table*}[ht]
    \centering
    \begin{tabular}{ccccccccccc}
        \toprule
        \multirow{2}{*}{methods} & \multicolumn{5}{c}{\textit{UCF101}} & \multicolumn{5}{c}{\textit{Caltech101}} \\ \cmidrule(lr){2-6} \cmidrule(lr){7-11}
         & 0.0 & 0.2 & 0.4 & 0.6 & 0.8 & 0.0 & 0.2 & 0.4 & 0.6 & 0.8 \\ \midrule
        CLIP-Adapter & 73.43 & 70.16 & 64.55 & 60.14 & 55.64 & 91.81 & 89.82 & 88.11 & 85.15 & 77.81 \\ 
        (CLIP-Adapter)+CRoF & 74.89$^{\uparrow}$ & 69.31$^{\downarrow}$ & 65.74$^{\uparrow}$ & 65.9$^{\uparrow}$ & 58.13$^{\uparrow}$ & 91.36$^{\downarrow}$ & 90.47$^{\uparrow}$ & 88.97$^{\uparrow}$ & 89.05$^{\uparrow}$ & 84.14$^{\uparrow}$ \\ 
        Tip-Adapter-F & 75.5 & 65.98 & 55.83 & 40.07 & 18.93 & 91.89 & 89.78 & 79.8 & 58.3 & 26.53 \\ 
        (Tip-Adapter-F)+CRoF & 75.31$^{\downarrow}$ & 70.92$^{\uparrow}$ & 63.65$^{\uparrow}$ & 61.49$^{\uparrow}$ & 54.03$^{\uparrow}$ & 91.97$^{\uparrow}$ & 89.61$^{\downarrow}$ & 88.92$^{\uparrow}$ & 85.48$^{\uparrow}$ & 78.01$^{\uparrow}$ \\ 
        \bottomrule
    \end{tabular}
    \caption{Accuracy (\%) on asymmetric noise (10-shot).}
    \label{tab:asymmetric}
\end{table*}


We evaluate our method on symmetric noise and asymmetric noise.  Symmetric noise \cite{ren2018learning} randomly replaces partial labels with other labels by noise ratio $\delta$. 
Asymmetric noise \cite{Li2020DivideMix}  replaces the labels of each class with incorrect labels from the same class for stronger noise data.
$\delta \in (0,1)$, represents the proportion of noisy labels in each class.

\textbf{(2) Baselines and Evaluation}

To evaluate the effectiveness of CRoF and its plug-in attribute for CLIP-based models, we select two typical CLIP-based few-shot learning baselines: CLIP-Adapter \cite{gao2024clip} and Tip-Adapter-F \cite{zhang2022tip}. 
These two base models are classic works in fine-tuned CLIP-based few-shot learning and have inspired various subsequent studies.
To verify the influence of our TPG  prompt, we compared it with CLIP prompt and CaFo.
We also compare the robustness between CRoF and traditional few-shot learning listed in  \cref{sec:re1}.

\textbf{(3) Implementation Details}

For CLIP, ResNet-50 \cite{He_2016_CVPR} is the backbone for image feature extraction and transformer  \cite{NIPS2017_3f5ee243} for text encoding.
During fine-tuning, the residual parameter $\lambda$ is  0.2, the same as in CLIP-Adapter. AdamW optimization algorithm \cite{kingma2014adam} is used with a learning rate of $10^{-3}$ and a cosine scheduler.
 $\alpha$, $\beta$ and $\gamma$ are set to 0.8, 0.8 and 0.9 respectively, and \textit{top-3} is used in selecting the candidate set.
It runs on an NVIDIA RTX 3090 GPU (24GB) and PyTorch 1.12.1.

\begin{figure*}[t]
    \centering
    \begin{subfigure}{0.24\linewidth}
         \centering
         \includegraphics[width=\linewidth]{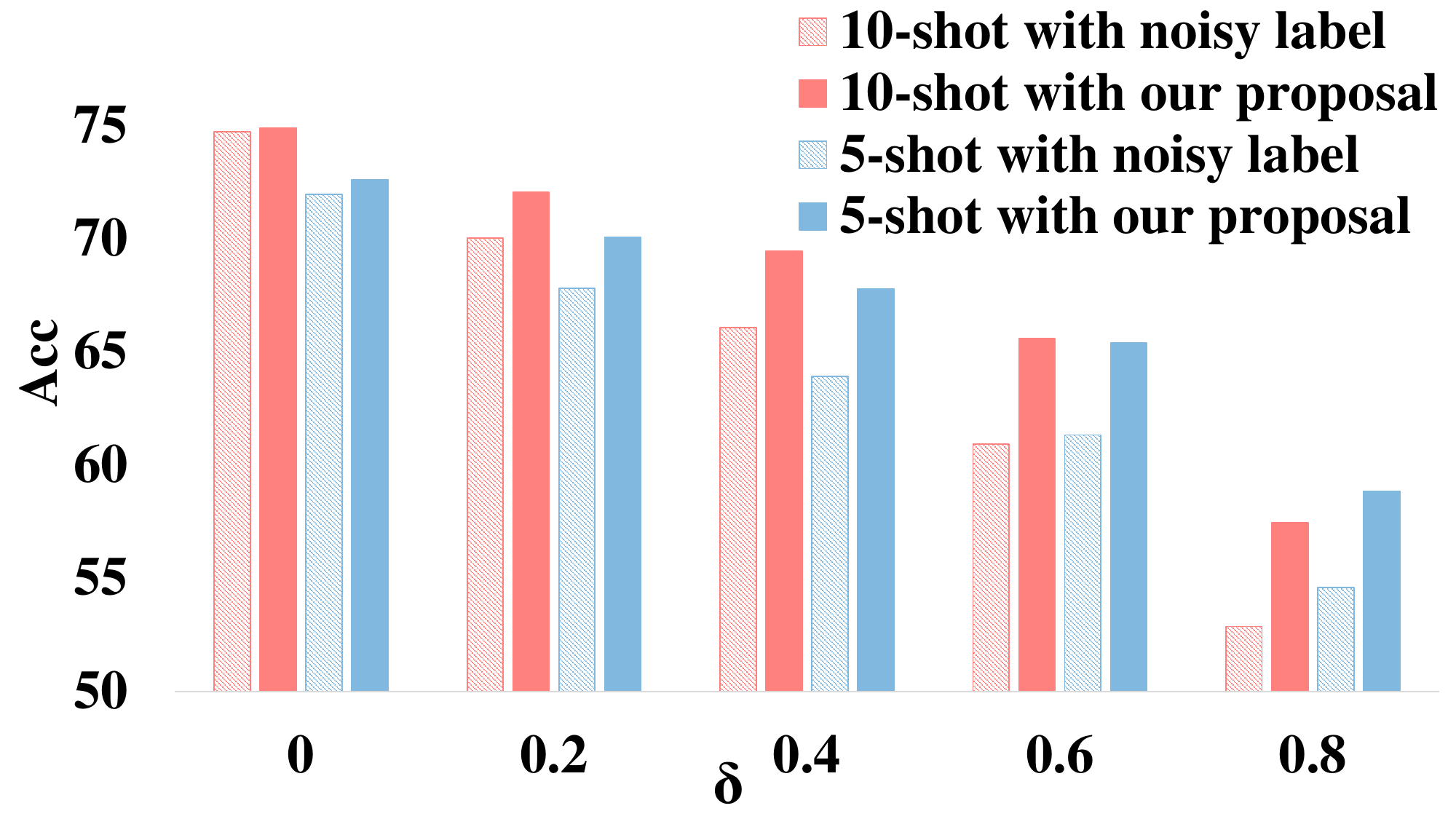}
         \caption{average on 7 datasets}
         \label{fig:average}
     \end{subfigure}
    \begin{subfigure}{0.24\linewidth}
        \centering
        \includegraphics[width=\linewidth]{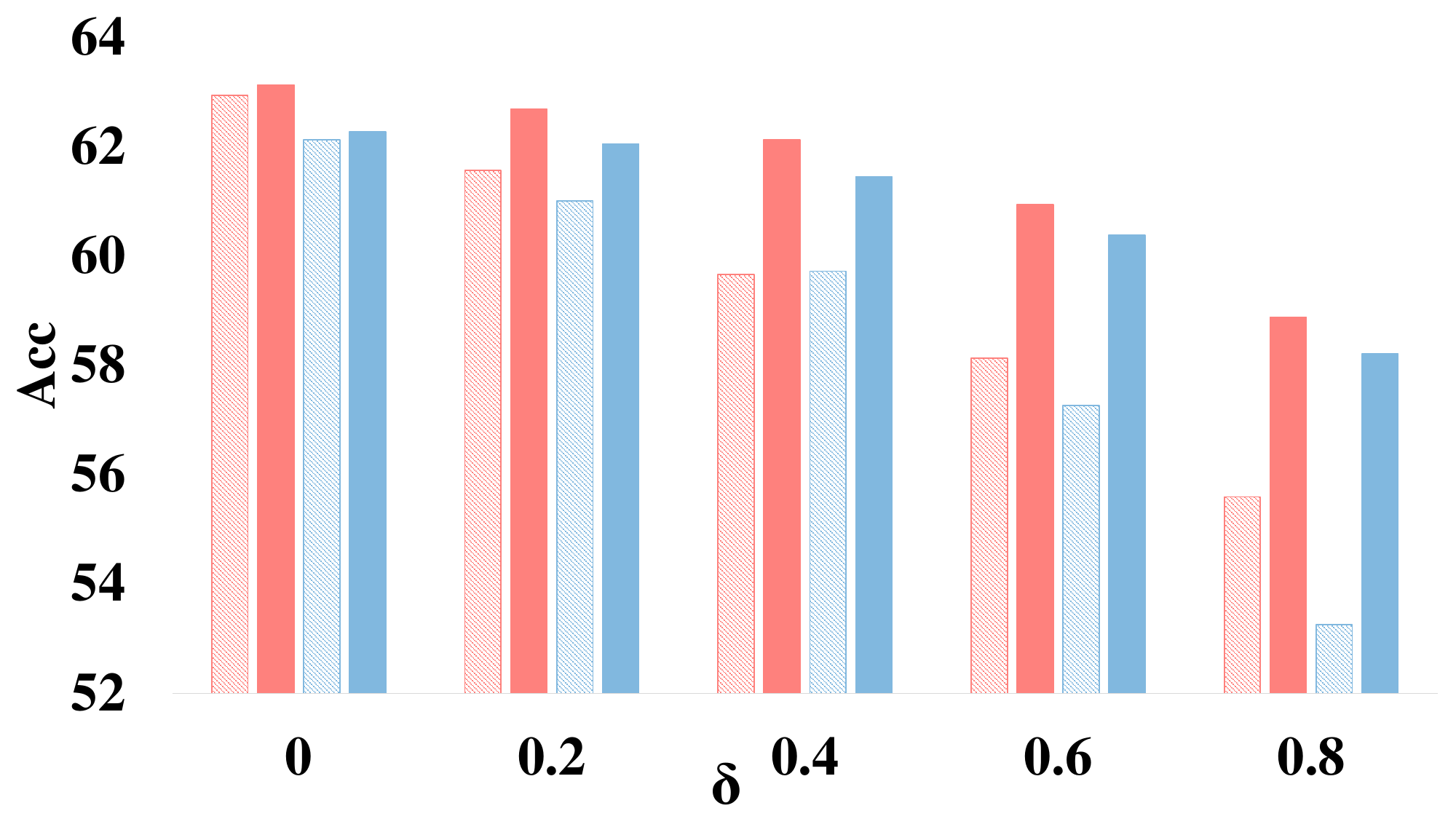}
        \caption{\textit{ImageNet}}
        \label{fig:imagenet}
    \end{subfigure}
    \begin{subfigure}{0.24\linewidth}
        \centering
        \includegraphics[width=\linewidth]{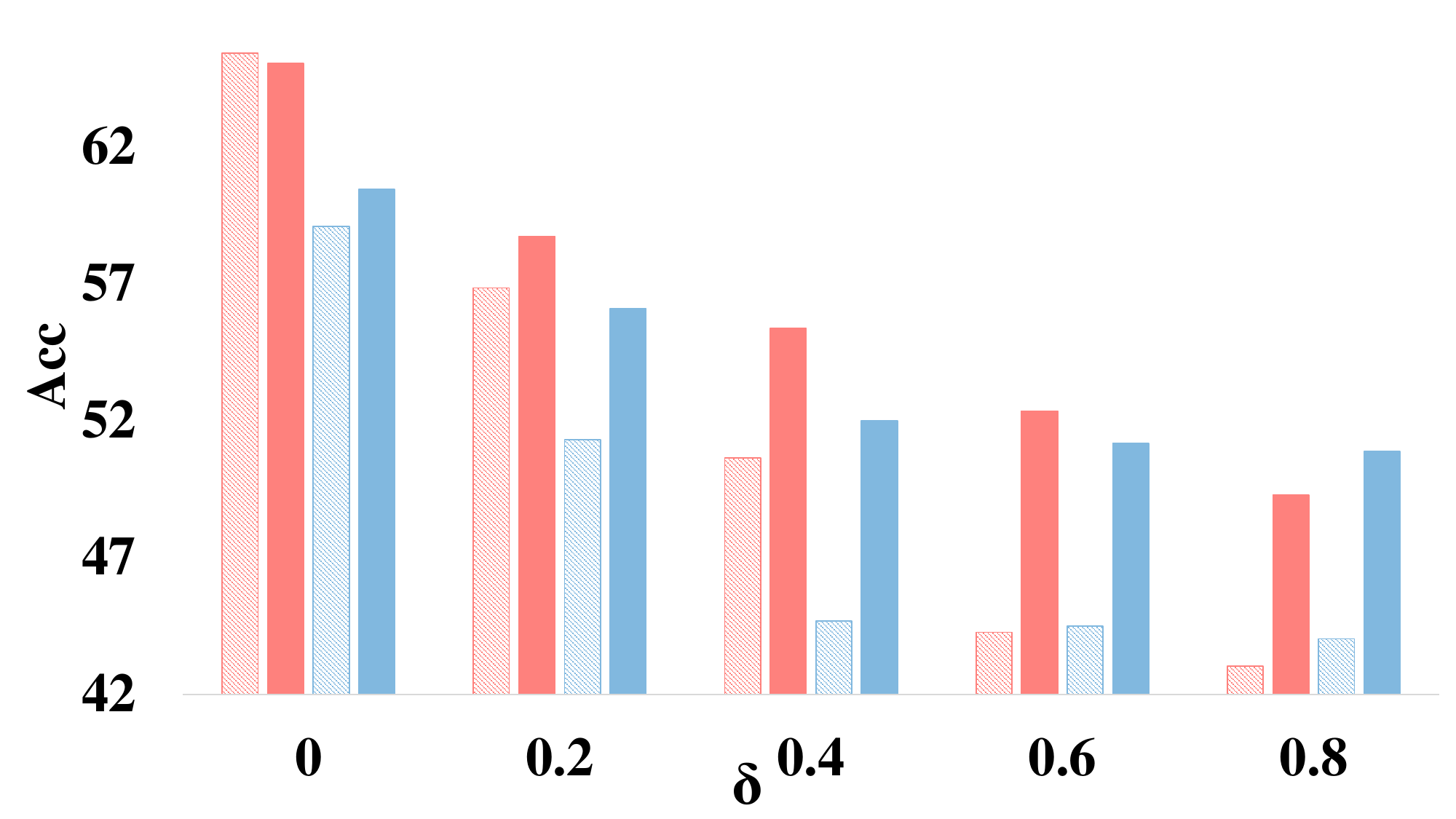}
        \caption{\textit{DTD}}
        \label{fig:dtd}
    \end{subfigure}    
    \begin{subfigure}{0.24\linewidth}
        \centering
        \includegraphics[width=\linewidth]{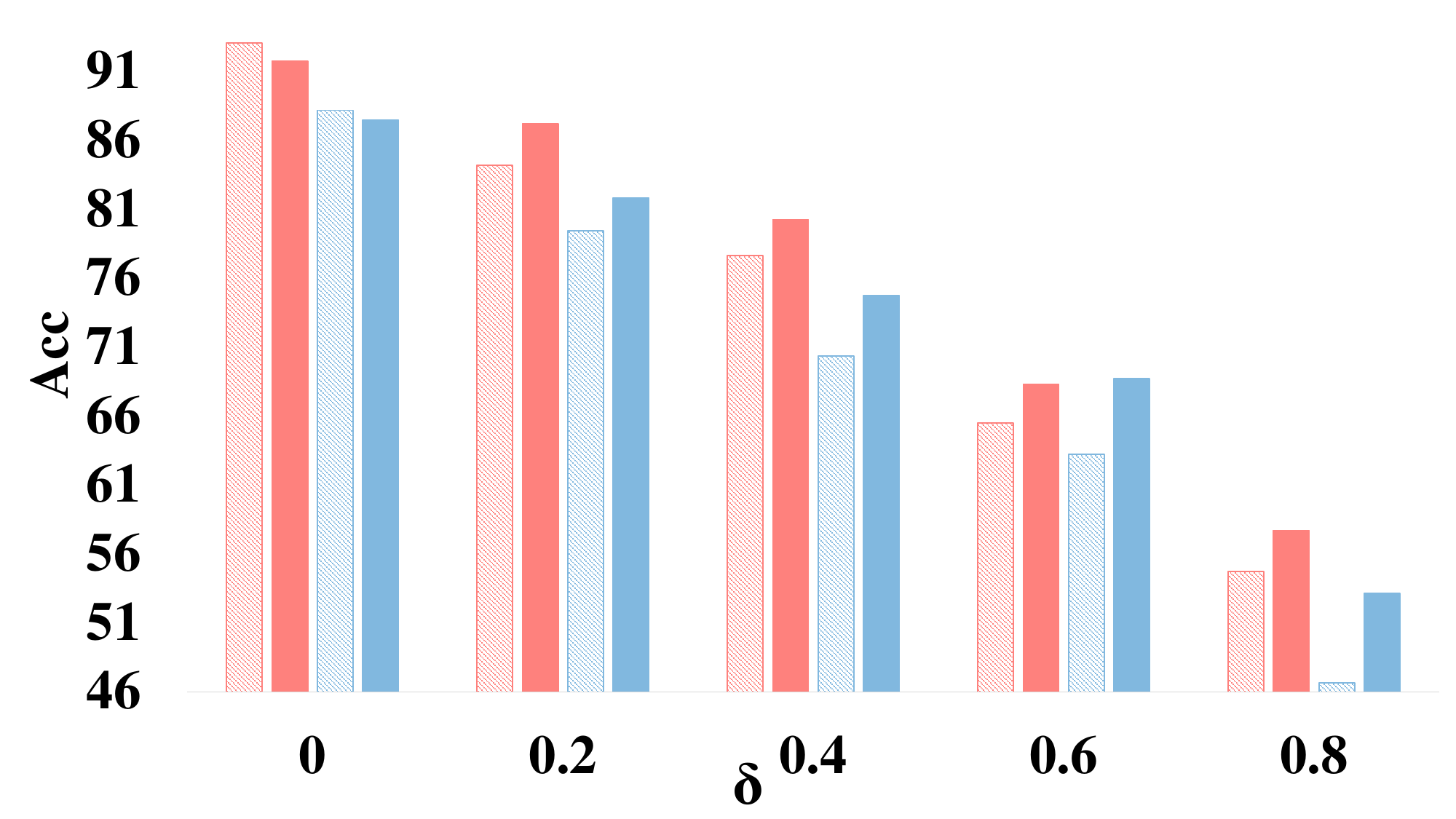}
        \caption{\textit{Flowers102}}
        \label{fig:Flowers102}
    \end{subfigure}   
    
    \begin{subfigure}{0.24\linewidth}
        \centering
        \includegraphics[width=\linewidth]{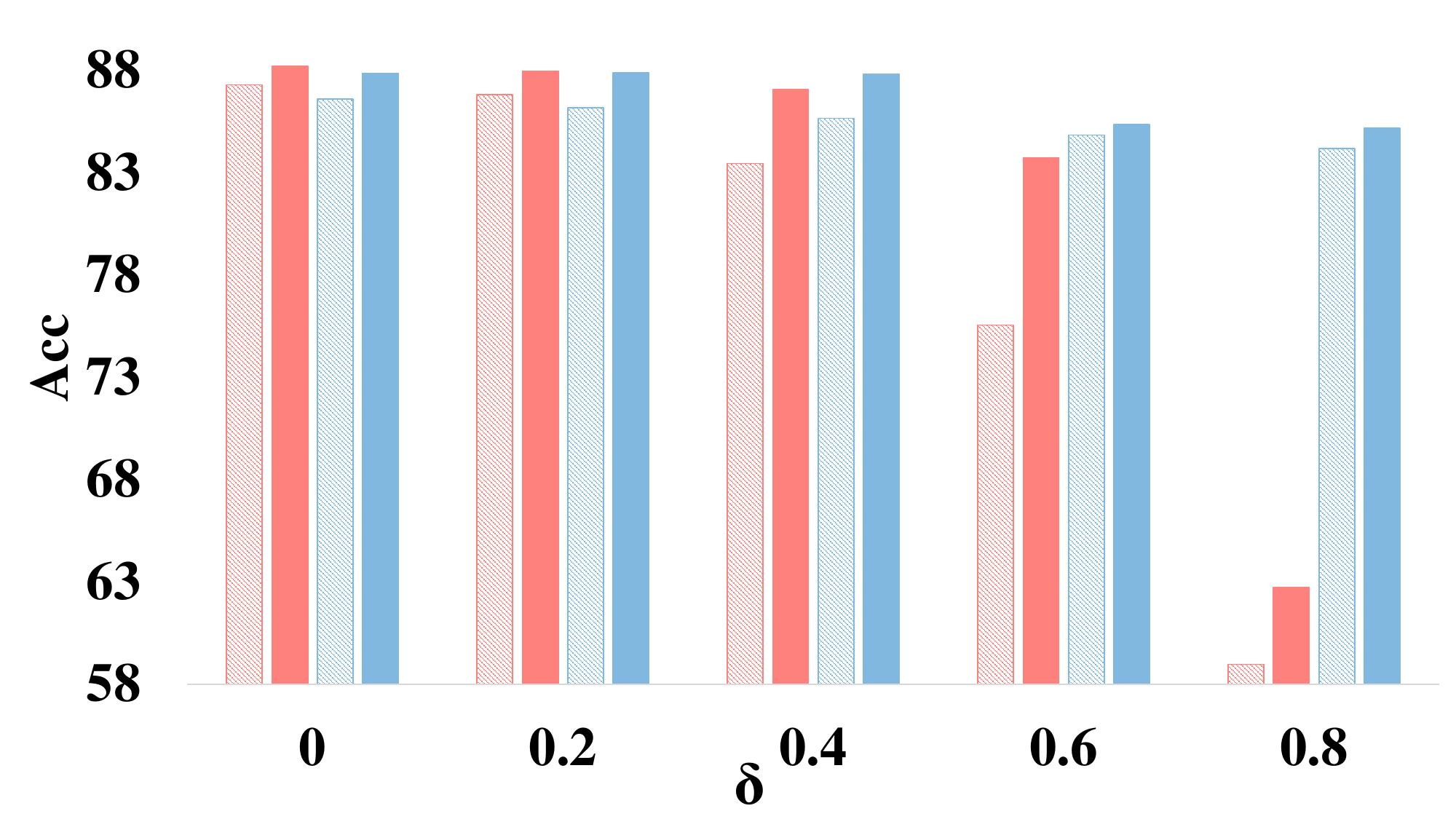}
        \caption{\textit{OxfordPets}}
        \label{fig:OxfordPets}
    \end{subfigure}
    \begin{subfigure}{0.24\linewidth}
        \centering
        \includegraphics[width=\linewidth]{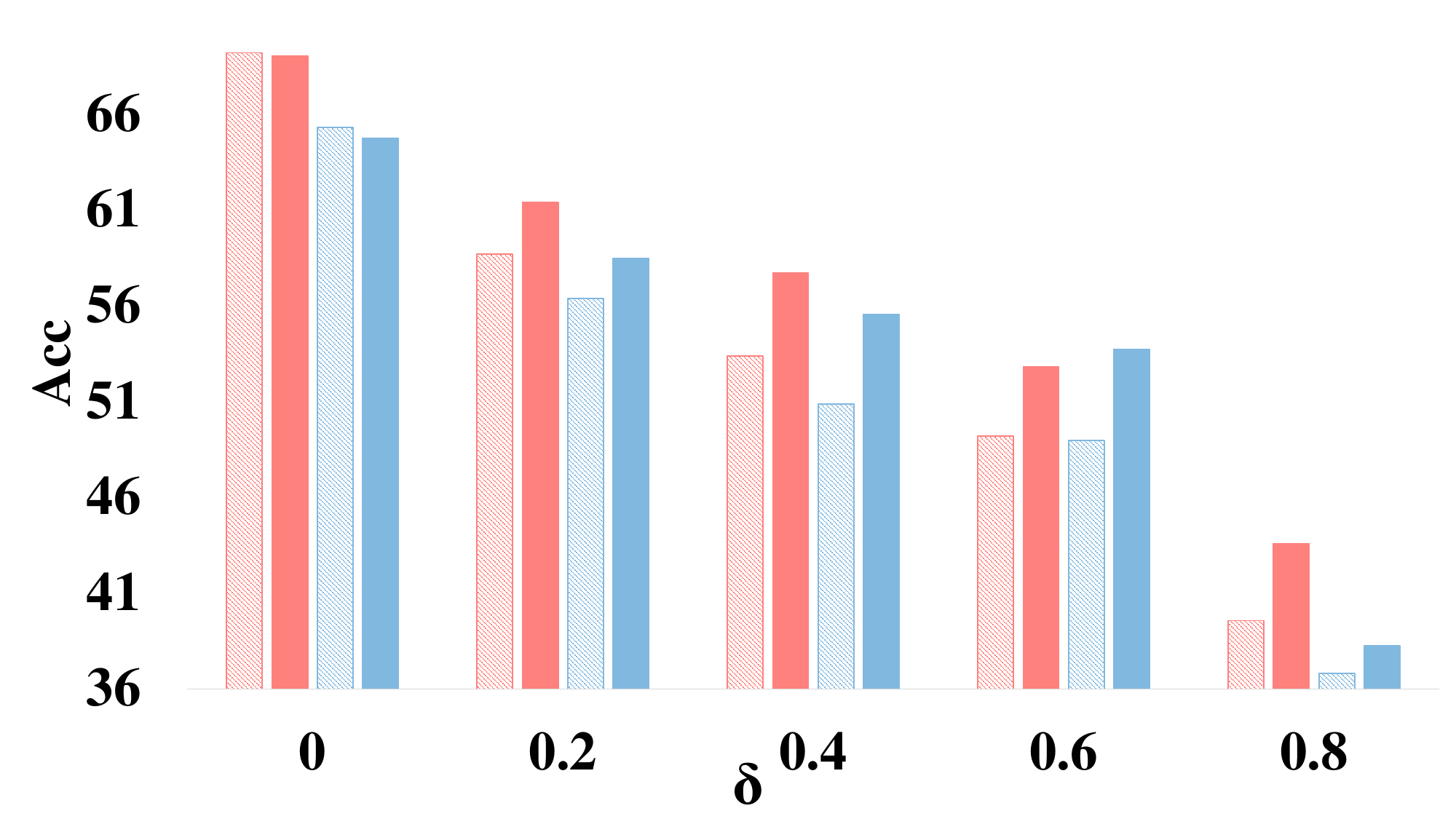}
        \caption{\textit{StanfordCars}}
        \label{fig:StanfordCars}
    \end{subfigure}        
    \begin{subfigure}{0.24\linewidth}
        \centering
        \includegraphics[width=\linewidth]{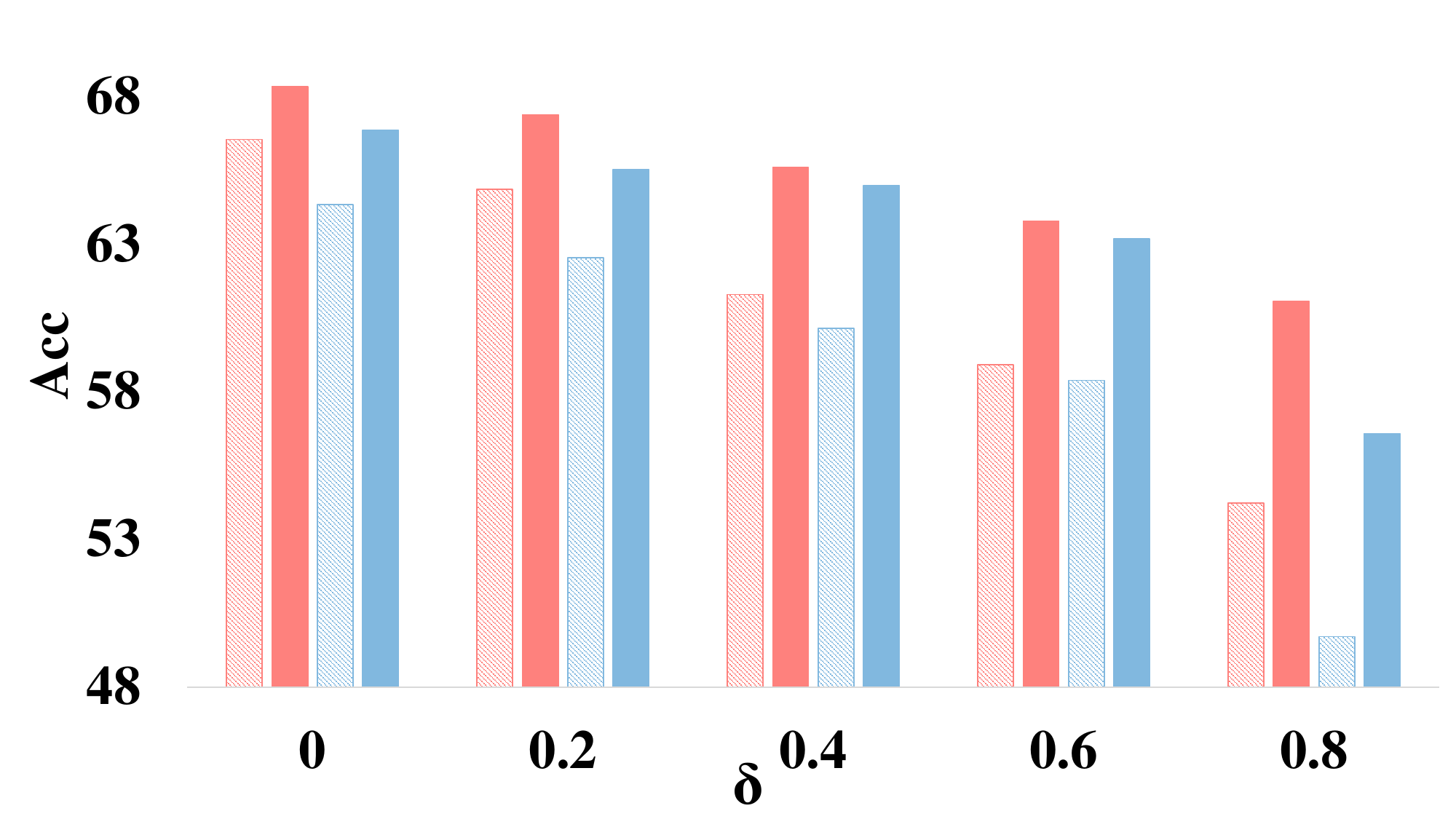}
        \caption{\textit{SUN397}}
        \label{fig:sun397}
    \end{subfigure}    
    \begin{subfigure}{0.24\linewidth}
        \centering
        \includegraphics[width=\linewidth]{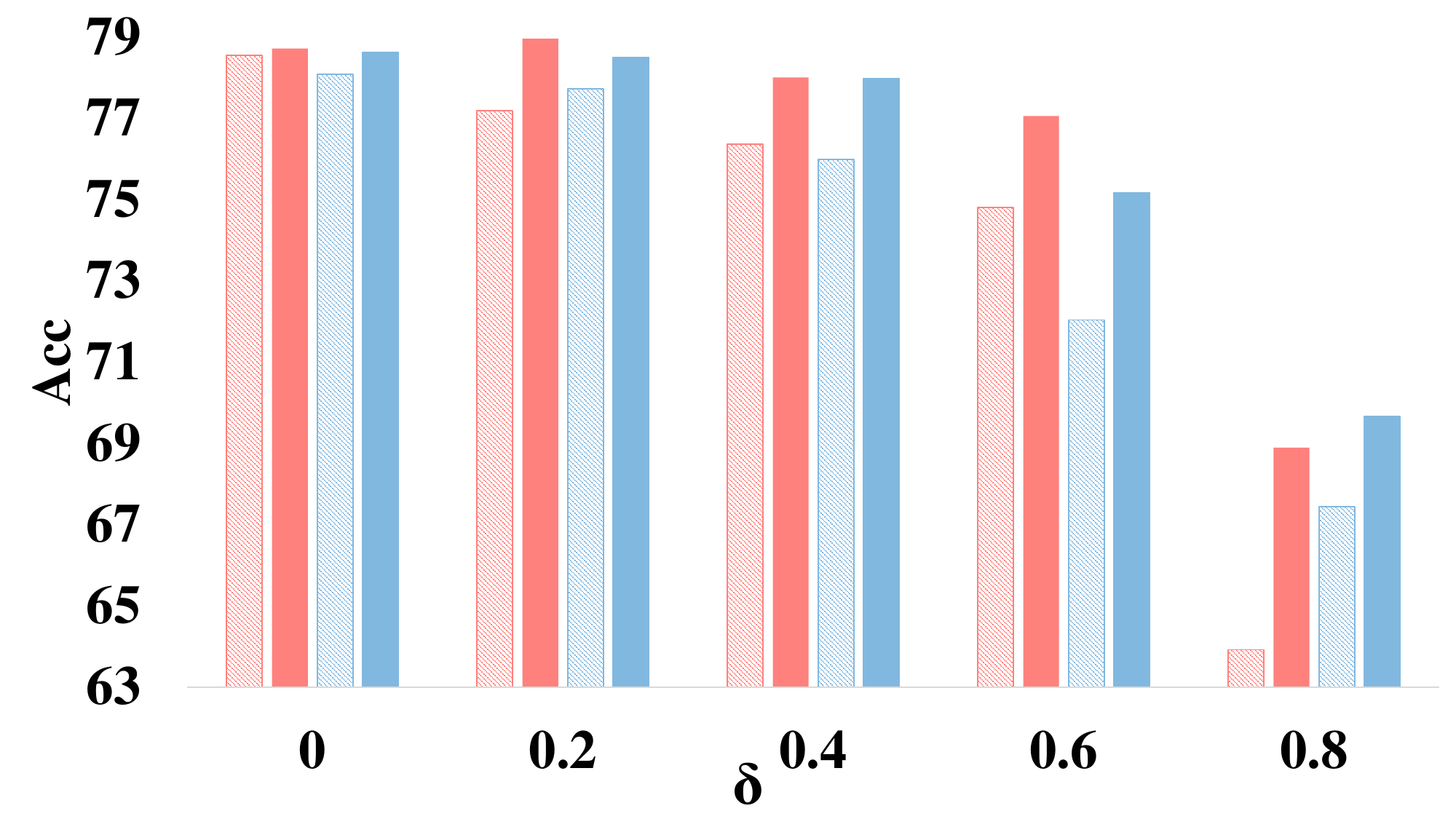}
        \caption{\textit{Food101}}
        \label{fig:Food101}
    \end{subfigure}

    \caption{Accuracy (\%) on other typical datasets in 5-shot and 10-shot with different symmetric noise ratios.}
    \label{fig:full}
\end{figure*}

\begin{figure*}[t]
    \centering
    \begin{subfigure}{0.22\linewidth} 
         \centering
         \includegraphics[width=\linewidth, height=0.8\linewidth]{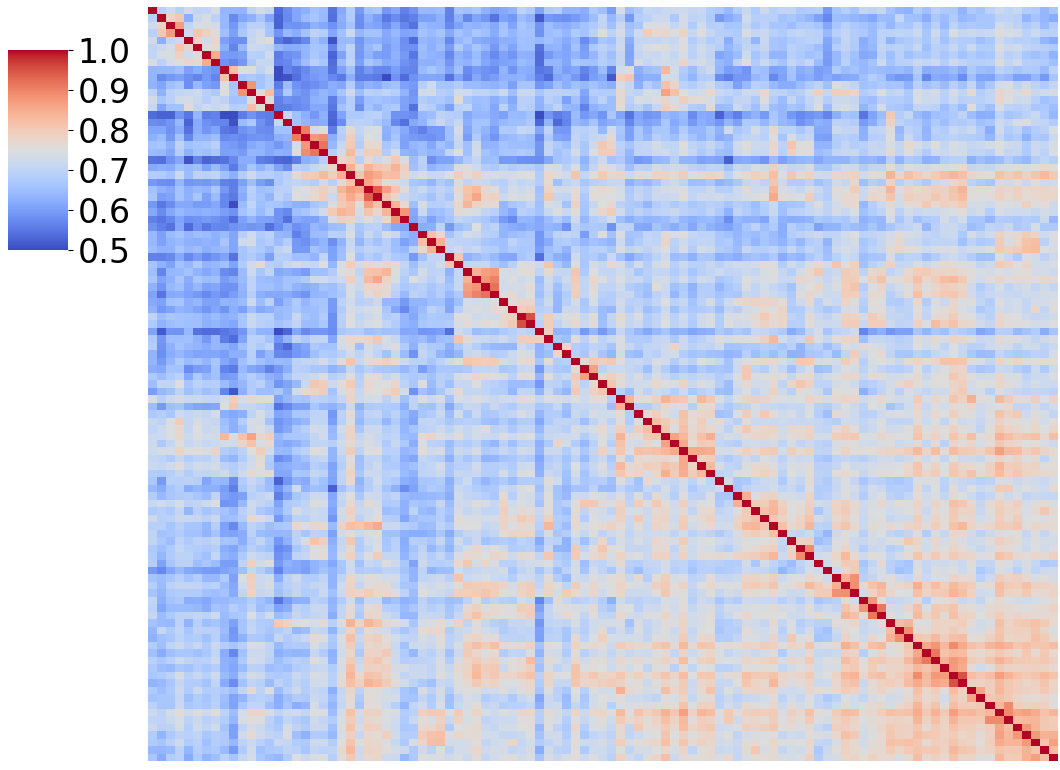}
         \caption{Org on \textit{UCF101}}
         \label{fig:ucforg}
     \end{subfigure}
     \hspace{0.001\linewidth} 
    \begin{subfigure}{0.22\linewidth}
        \centering
        \includegraphics[width=\linewidth, height=0.8\linewidth]{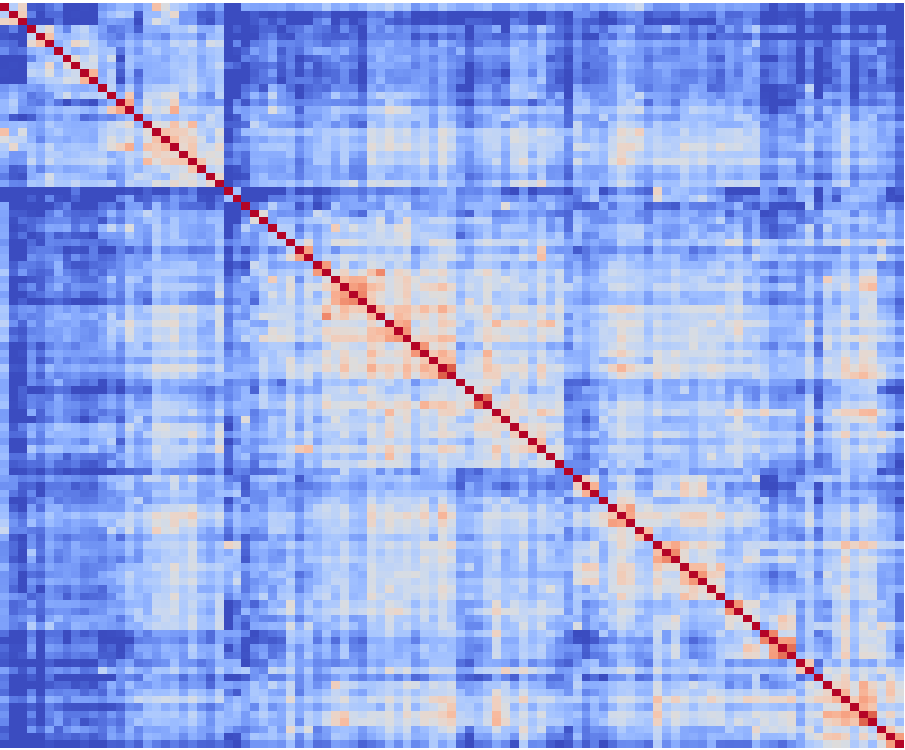}
        \caption{CaFo on \textit{UCF101}}
        \label{fig:ucfcafo}
    \end{subfigure}
    \hspace{0.001\linewidth} 
    \begin{subfigure}{0.22\linewidth}
        \centering
        \includegraphics[width=\linewidth, height=0.8\linewidth]{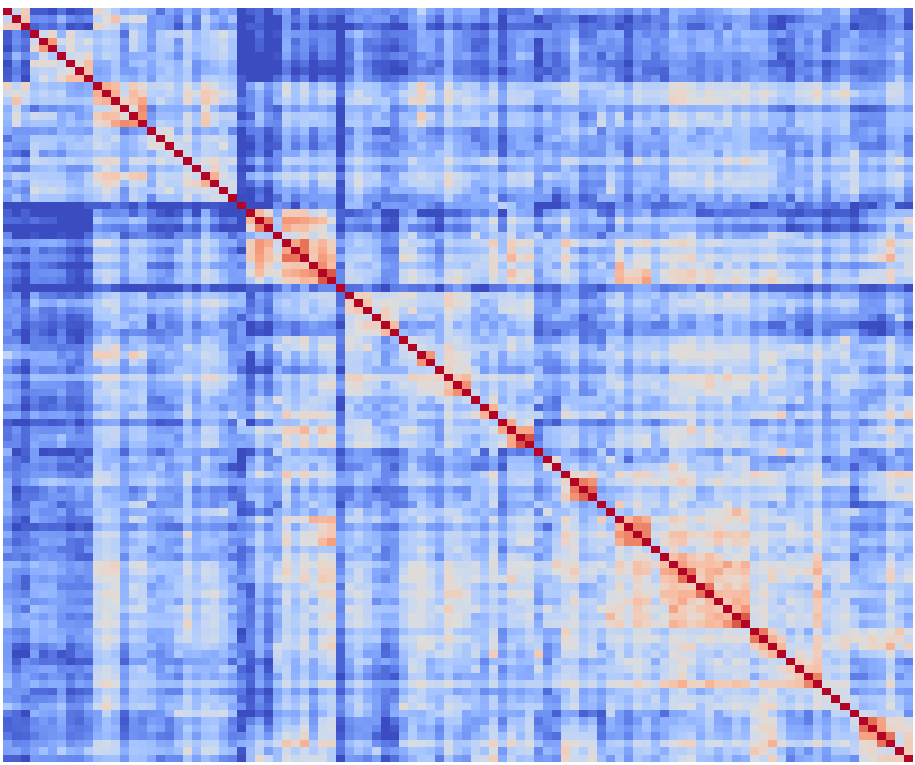}
        \caption{Sup on \textit{UCF101}}
        \label{fig:ucfmine}
    \end{subfigure}    
    \hspace{0.001\linewidth} 
    \begin{subfigure}{0.22\linewidth}
        \centering
        \includegraphics[width=\linewidth, height=0.8\linewidth]{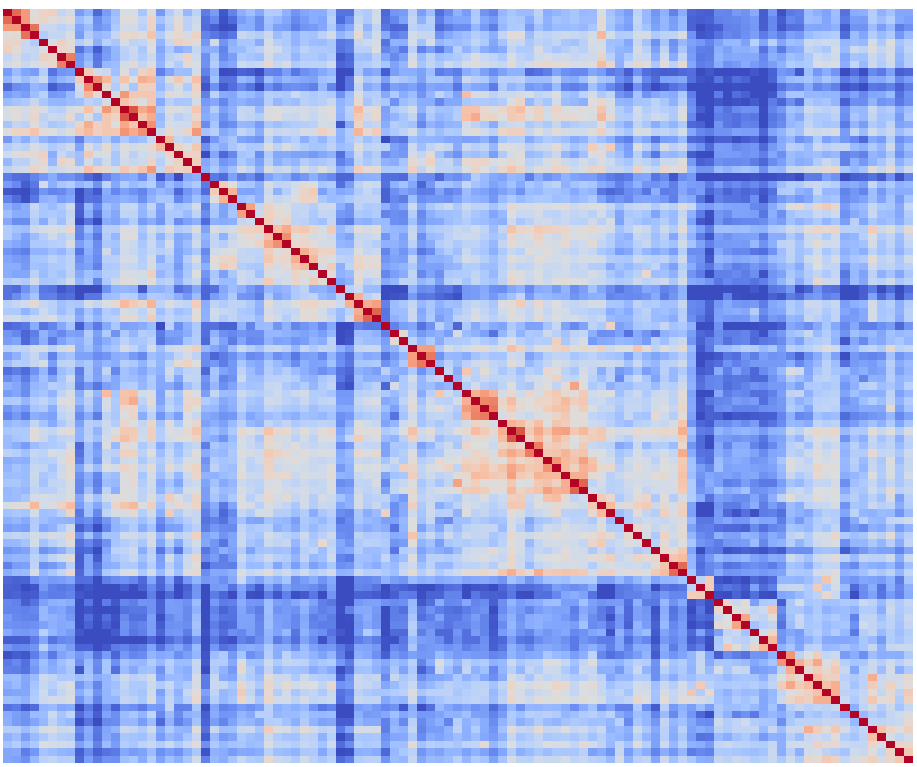}
        \caption{TPG on \textit{UCF101}}
        \label{fig:ucfhunhe}
    \end{subfigure}   
    
    \vspace{0.1cm} 

    \begin{subfigure}{0.22\linewidth}
        \centering
        \includegraphics[width=\linewidth, height=0.8\linewidth]{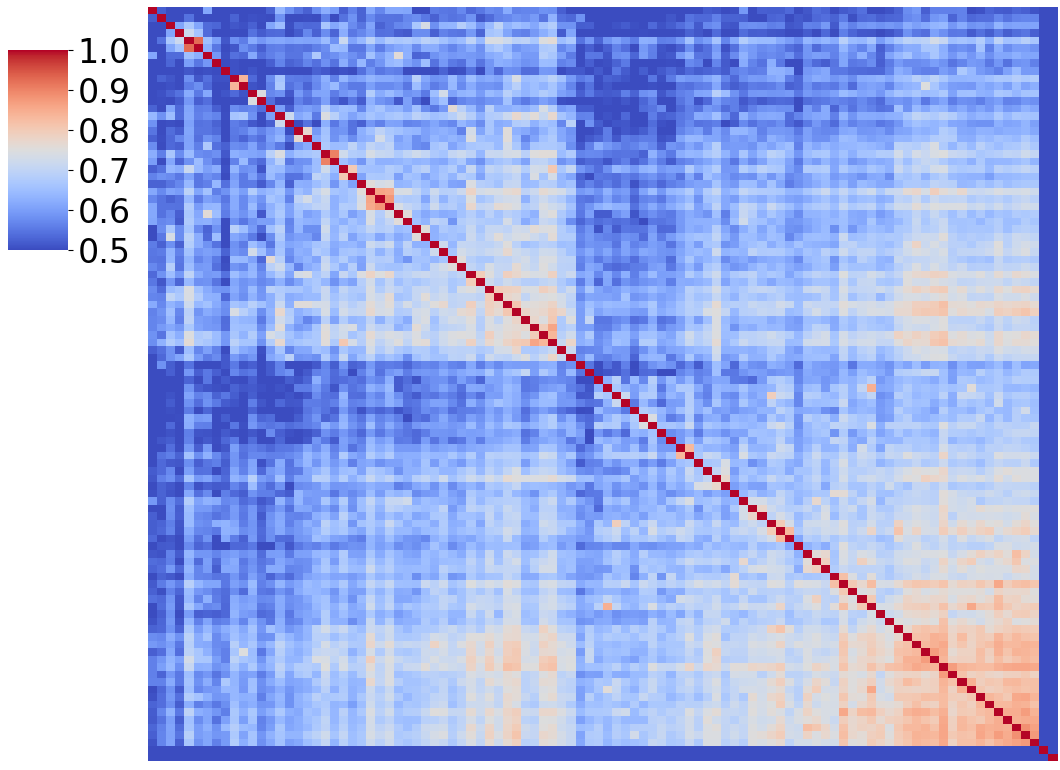}
        \caption{Org on \textit{Caltech101}}
        \label{fig:calorg}
    \end{subfigure}
    \hspace{0.001\linewidth} 
    \begin{subfigure}{0.22\linewidth}
        \centering
        \includegraphics[width=\linewidth, height=0.8\linewidth]{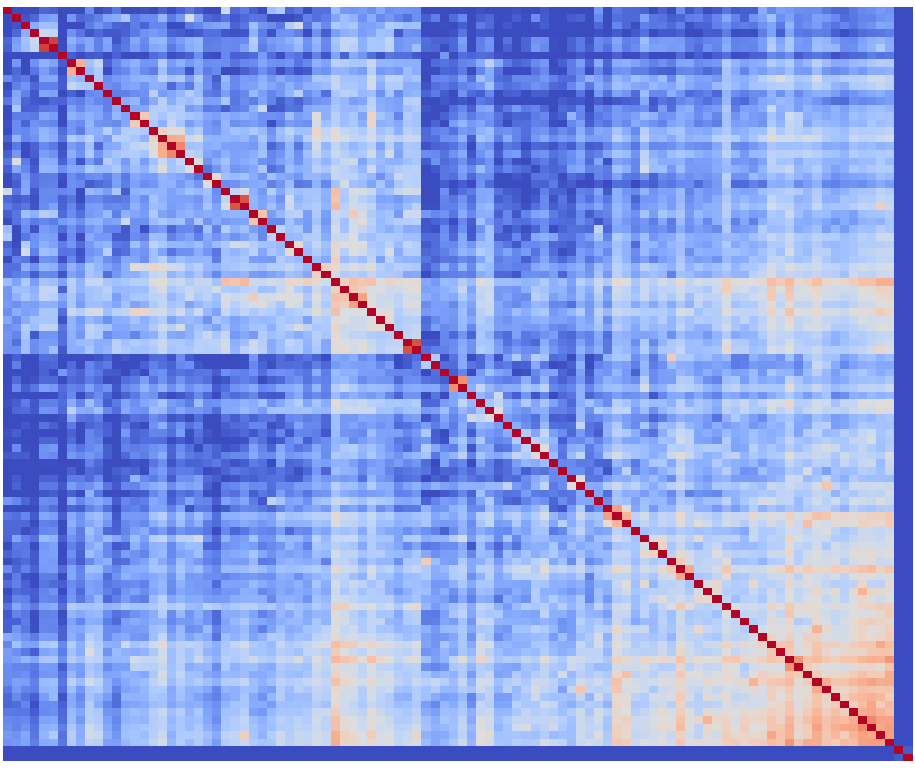}
        \caption{CaFo on \textit{Caltech101}}
        \label{fig:calcafo}
    \end{subfigure}        
    \hspace{0.001\linewidth} 
    \begin{subfigure}{0.22\linewidth}
        \centering
        \includegraphics[width=\linewidth, height=0.8\linewidth]{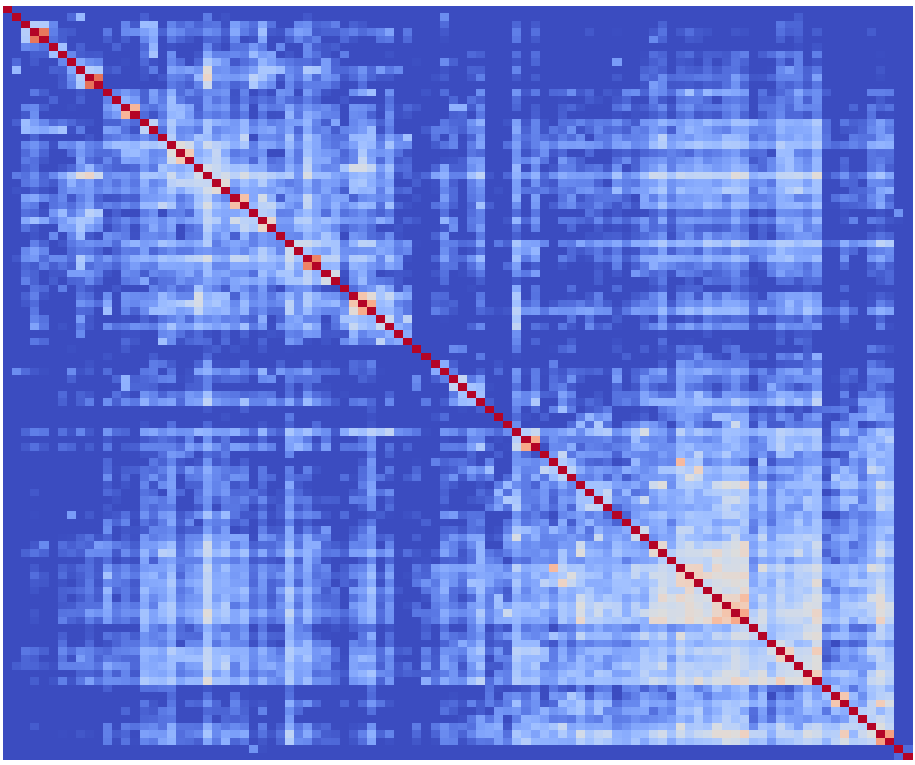}
        \caption{Sup on \textit{Caltech101}}
        \label{fig:calmine}
    \end{subfigure}    
    \hspace{0.001\linewidth} 
    \begin{subfigure}{0.22\linewidth}
        \centering
        \includegraphics[width=\linewidth, height=0.8\linewidth]{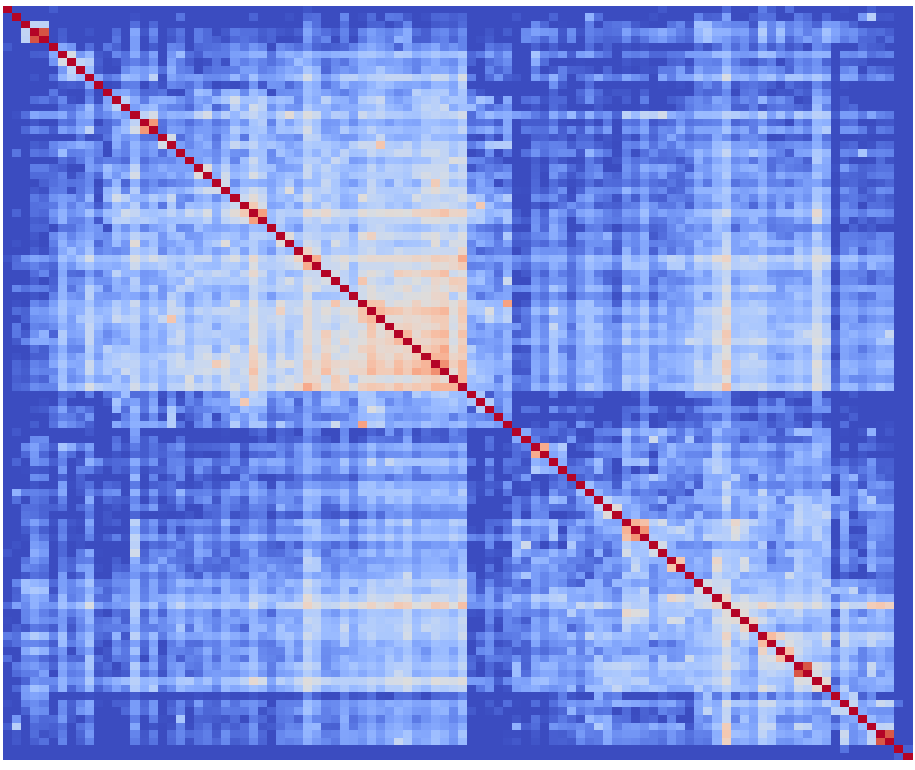}
        \caption{TPG on \textit{Caltech101}}
        \label{fig:calhunhe}
    \end{subfigure}

    \caption{Heatmap on the similarity among the category embedding with different prompts. Org is the basic prompt, CaFo is prompt in \cite{Zhang_2023_CVPR}, Sup is our supplement prompt, and TPG is the noimalization of Sup and CaFo. }
    \label{fig:relitu}
\end{figure*}

\subsection{Experimental Results}\label{sec:experrelt}

To validate the robustness of our proposed CRoF as a plug-in for CLIP-based models  on noisy labels,
we compare the performance of the base models (CLIP-Adapter and  Tip-Adapter-F) and traditional few-shot learning models.
As shown in \cref{tab:symmetric} and \cref{tab:asymmetric}, upward and downward arrows indicate performance increases and decreases, respectively.

\textbf{(1) Results on Symmetric Noise}


Overall, CRoF enhances the base models by handling noisy label issues.
On both datasets, there is an obvious and significant improvement for both models in \cref{tab:symmetric}.
The maximal average increase is approximately  10\% in \textit{Caltech101} with (Tip-Adapter-F)+CRoF  across  all $\delta$ in 10-shot. 
Besides, CRoF would not affect the comparative relationship between the base models.
Out of 16 comparisons for each model, CLIP-Adapter outperforms Tip-Adapter-F in 11 cases. 
After adding CRoF, 15 comparisons still favor CLIP-Adapter, demonstrating its robust performance.

CRoF is not sensitive to noisy data.
As the noise ratio $\delta$ increases, it is not surprising that the accuracy has reduction.
However, the gap between the lower $\delta$ and higher $\delta$ with CRoF is smaller than that of base models.
For example in \textit{Caltech101}, there is a decrease of 26.74\%  between 89.13\% ($\delta=0.2$) and 62.39\% ($\delta=0.8$) in  Tip-Adapter-F on 5-shot. 
The decrease of Tip-Adapter-F with plug-in CRoF is only 9.66\%, which is far lower than 26.74\%.

In the cleaning dataset, the accuracy of vanilla CLIP is 61.35\% and 85.92\% in \textit{UCF101} and \textit{Caltech101} respectively in zero-shot learning. 
All fine-tuned models outperform vanilla CLIP, which verifies the effectiveness of domain adaption.
CRoF also gives satisfying classification in most cases due to its ability to correct the noisy correspondence from CLIP.

\textbf{(2) Results on Asymmetric Noise}

The evaluations in a more extreme noise condition are illustrated in \cref{tab:asymmetric}.
Compared to the results in \cref{tab:symmetric}, most models exhibit lower accuracy.
However, the base models with CRoF  have superior performance.
Although catch-tuning Tip-Adapter-F performs worst, especially when ($\delta>0.5$) due to the majority of incorrect labels with the same class, our CRoF can still refine Tip-Adapter-F.


\textbf{(3) Comparison with Traditional Few-shot Learning}

CRoF shows a dramatic improvement in \cref{tab:comparisonfew}, which has two advantages.
For one thing, CRoF is a fine-tuned model on pre-trained CLIP and the prior knowledge is more friendly for noisy few-shot tasks.
Traditional few-shot learning is trained on a close set with an upper bound.
For another, sufficient prior knowledge  makes CRoF adapt to more classes, 20 classes in \textit{miniImageNet} and 160 in \textit{tieredImageNet}, which are the total number of classes in the testing of traditional FSL.
However, the traditional noisy FSL only follows the paradigm of ($5$-way, $5$-shot) due to its learning mechanism.

\begin{table}[ht]
    \centering
    
    \begin{tabular}{llcc|cc}
        \toprule
        \multirow{2}{*}{Methods} & \multirow{2}{*}{} & \multicolumn{2}{c}{\textit{miniImageNet} } & \multicolumn{2}{c}{\textit{tieredImageNet}} \\
        \cmidrule(lr){3-4} \cmidrule(lr){5-6}
        & & 0.2 & 0.4 & 0.2 & 0.4 \\
        \midrule
        Proto Nets \cite{snell2017prototypical} & & 58.40 & 46.04 & 55.20 & 44.14 \\
        Matching Nets \cite{vinyals2016matching} & & 56.55 & 44.65 & 53.33 & 44.72 \\
        Relation Nets \cite{sung2018learning} & & 57.06 & 45.46 & 55.09 & 44.73 \\
        MAML \cite{finn2017model} & & 55.77 & 45.01 & 54.19 & 41.91 \\
        \midrule
        RNNP  \cite{Mazumder_2021_WACV} & & 59.97 & 46.78 & 58.47 & 46.57 \\
        RapNet  \cite{9072304} & & 59.01 & 43.91 & 56.28 & 42.50 \\
        APPN \cite{chen2024appn} & & 62.68 & 50.41 & 59.01 & 46.28 \\
        TraNFS  \cite{Liang_2022_CVPR} & & 65.08 & 56.65 &  67.67 & 58.88 \\
        \midrule
        CRoF & & \textbf{96.07} & \textbf{95.21} & \textbf{68.52} & \textbf{68.05} \\
        \bottomrule
    \end{tabular}
    \caption{Comparison with traditional few-shot paradigm.}
    \label{tab:comparisonfew}
\end{table}

\begin{table}[htbp]
\centering
\begin{tabular}{ccccccc}
\toprule
No.&TPG & FT& WT & 5-shot & 10-shot \\
\midrule
1&\texttimes & \texttimes & \texttimes & 61.56 & 60.32 \\
2&\texttimes & \checkmark & \texttimes & 64.08 & 65.77 \\
3&\texttimes & \texttimes & \checkmark & 64.76 & 64.31 \\
4&\texttimes & \checkmark & \checkmark & 66.22 & 67.72 \\
5&\checkmark & \texttimes & \texttimes & 65.03 & 64.68 \\
6&\checkmark & \checkmark & \texttimes & 66.64 & 66.88 \\
7&\checkmark & \texttimes & \checkmark & 67.22 & 67.59 \\
8&\checkmark & \checkmark & \checkmark & \textbf{67.59} & \textbf{69.87} \\
\bottomrule
\end{tabular}
\caption{Accuracy (\%) of CLIP-Adapter with CRoF in \textit{UCF101}.}
\label{tab:ablation}
\end{table}

\subsection{Genelization on other Datasets}
 To evaluate the generalization of CRoF, we conduct experiments by using  CLIP-Adapter and (CLIP-Adapter)+CRoF  on seven additional typical datasets in  \cref{fig:full}.
 CRoF improves classification accuracy and the improvement is more obvious as  $\delta$ increases.
 On \textit{ImageNet}, the improvement of CRoF is minimal. This is because the zero-shot accuracy of CLIP is 60.32\%, and only grows to 67.72\% after fine-tuning on the clean 10-shot dataset. 
 Vanilla CLIP is already highly powerful to \textit{ImageNet}, thus fine-tuning has less impact.

\subsection{Ablation Study}\label{sec:main ab}

\textbf{(1) Module Ablation}

We demonstrate the superior performance achieved through the combined effectiveness of our there proposed modules at $\delta$=0.4 in \cref{tab:ablation}.
TPG,  FT, WT represent task-oriented prompt generator, fine-tuning CLIP, and multiple label weighting respectively.
Every single module has improvement compared to vanilla CLIP (No.1 row), where TPG and FT perform best in 5-shot and 10-shot respectively.
Any combination of the two modules also achieves performance gains with FT+WT achieving the highest increase in 10-shot.

\textbf{(2) Prompt Analysis}

To assess the impact of the task-oriented prompt generator, we visualize the similarity among the category embeddings in the heatmap in \cref{fig:relitu}.
TPG provides the appropriate inter-class distances, enhancing category embedding differentiation and preserving text-image prior knowledge invariance. 
Furthermore, we applied four different prompts to the CLIP-Adapter, as shown in  \cref{tab:prompt}.
TPG prompts consistently exhibit the best performance.
It demonstrates the resilience of our designed prompts against noisy labels. 

\begin{table}[ht]
    \centering
    \setlength{\tabcolsep}{0.6mm}
    \resizebox{\linewidth}{!}{ 
    \begin{tabular}{ccccccccccc}
        \toprule
        \multirow{2}{*}{methods} & \multicolumn{5}{c}{\textit{Caltech101}} & \multicolumn{5}{c}{\textit{UCF101}} \\ \cmidrule(lr){2-6} \cmidrule(lr){7-11}
         & 0.0 & 0.2 & 0.4 & 0.6 & 0.8 & 0.0 & 0.2 & 0.4 & 0.6 & 0.8 \\ \midrule
        Org & \textbf{91.8} & 90.4 & 90.2 & 88.3 & 82.4 & 73.4 & 66.9 & 63.7 & 61.0 & 52.4 \\ 
        CaFo & 91.3 & 90.5 & 89.6 & 88.2 & 81.7 & 73.8 & 67.8 & 65.1 & 62.5 & 53.4 \\ 
        Sup & 91.6 & 90.3 & 89.7 & 88.0 & 83.3 & \textbf{74.9} & 68.2 & 64.8 & 61.8 & 55.6 \\ 
        TPG & \textbf{91.8} & \textbf{90.6} & \textbf{90.2} & \textbf{88.9} & \textbf{83.9} & 74.7 & \textbf{68.7} & \textbf{67.1} & \textbf{63.2} & \textbf{55.9} \\ 
        \bottomrule
    \end{tabular}
    }
    \caption{Accuracy (\%) with different prompts.}
    \label{tab:prompt}
\end{table}

\subsection{Hyper-parameter Discussion}
This section discusses the roles of $\alpha$, $\beta$ and the number of epochs in the training process on \textit{UCF101}.
We conduct experiments in the 10-shot setting with symmetric noise in (CLIP-Adapter)+CRoF and CLIP-Adapter.


\textbf{(1) Loyalty to Original Label $\boldsymbol\alpha$}

\begin{figure}[ht]
  \centering
  \includegraphics[scale=0.79]{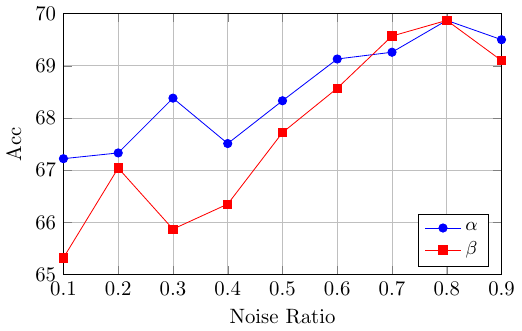}
  \caption{Accuracy (\%) with different $\alpha$ and $\beta$.}
    \label{fig:xiaorong}
\end{figure}

\cref{fig:xiaorong} shows that as $\alpha$ increases and $\beta$=0.8, the accuracy has two peaks at 0.3 and 0.8 respectively, and the maximum accuracy 69.87\% is reached at $\alpha$=0.8. The reason is that both the original label and the label with the highest similarity play a role. 
As $\alpha$ increases, the original label’s contribution becomes more significant, while the importance of the most similar label decreases.
At these two peaks,  the corrected label distribution and the original label play the dominant role respectively. 
This result validates our choice of parameter $\alpha$, indicating that greater weight should be assigned to the original labels to achieve optimal balance.

\textbf{(2) Confidence in the Most Similar Label $\boldsymbol{\beta}$}

The results with different $\beta$ and $\alpha$=0.8 are shown in \cref{fig:xiaorong}. When $\beta$ is 0.8, the accuracy is the highest, and as $\beta$ increases or decreases, the accuracy shows a decreasing trend.
It verifies our hypothesis that the label with the highest similarity should be given a larger weight when taking the multiple label weights strategy.

 \begin{figure}[ht]
  \centering
  \includegraphics[width=0.95\columnwidth]{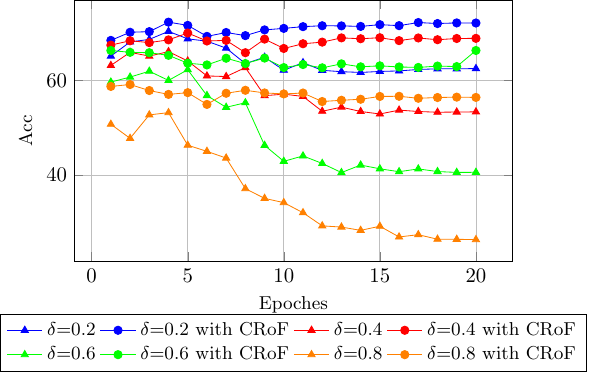}
  \caption{Accuracy (\%) during training with different $\delta$.}
    \label{fig:data_curves_marker_styles}
\end{figure}

\textbf{(3) Performance on Different Epochs}

To verify the performance of CRoF over different epochs, we set varying noise ratios and record the accuracy at different epochs, as illustrated in \cref{fig:data_curves_marker_styles}. 
The accuracy trend is different from traditional accuracy curves.
The main reason is that the few-shot adaptation is on the noisy dataset.
As is demonstrated in \cref{sec:experrelt} and \cref{sec:main ab}, fine-tuning is necessary on the limited trainable samples. 
However, we cannot give full confidence to the annotated labels.
If they are noisy labels for some reason, fine-tuned CLIP would lead to misclassification.
CRoF provides a safeguard to  CLIP-based models for robust few-shot tasks.
When the accuracy of fine-tuned CLIP starts to decline due to overfitting to noise, CRoF still has a stable classification performance.
It demonstrates that our method mitigates the negative impact of noise.


%% file: sec/5_conclusion.tex
\section{Conclusion}

We revisit noisy few-shot learning (FSL) tasks on the multiple-modal large-scale vision-language model, CLIP.
To enhance robustness in CLIP-based models, we propose a plug-in solution, CRoF.
CRoF aims to address the issue of noisy labels and noisy correspondence from CLIP through three optimizations.
Firstly, a task-oriented prompt generator increases the distance of the similar label in the textual embedding to avoid consuming noisy labels.
Secondly, a fine-tuned CLIP is trained to improve the generalization for FSL tasks in new domains.
Finally, multiple label weighting strategy balances the  CLIP prior knowledge and the original label information.
In the comprehensive experiments, three optimization modules power each other and their combination shows an excellent performance in few-shot tasks with noisy labels, especially at high noise ratios.